\documentclass[11pt,letterpaper]{article}
\usepackage{cogsys}
\usepackage[T1]{fontenc}
\usepackage{times}
\usepackage[pdftex]{graphicx} 

\usepackage{algorithm}
\usepackage{algorithmic}
\usepackage{enumitem}

\usepackage{natbib}
\ShortHeadings{Enabling Morally Sensitive Robotic Clarification Requests}
              {R. B. \ Jackson and T.\ Williams}

\begin{document} 

\title{Enabling Morally Sensitive Robotic Clarification Requests}
 
\author{Ryan Blake Jackson}{rbjackso@mines.edu}
\author{Tom Williams}{twilliams@mines.edu}
\address{MIRRORLab, Dept. of Computer Science, Colorado School of Mines, 
         Golden, CO 80401 USA}
\vskip 0.2in
 
\begin{abstract}
The design of current natural language oriented robot architectures enables certain architectural components to circumvent moral reasoning capabilities. One example of this is reflexive generation of clarification requests as soon as referential ambiguity is detected in a human utterance. As shown in previous research, this can lead robots to (1) miscommunicate their moral dispositions and (2) weaken human perception or application of moral norms within their current context. We present a solution to these problems by performing moral reasoning on each potential disambiguation of an ambiguous human utterance and responding accordingly, rather than immediately and naively requesting clarification. We implement our solution in the DIARC robot architecture, which, to our knowledge, is the only current robot architecture with both moral reasoning and clarification request generation capabilities. We then evaluate our method with a human subjects experiment, the results of which indicate that our approach successfully ameliorates the two identified concerns. 
\end{abstract}

\section{Introduction} \label{intro}
To accommodate the tremendous diversity of communicative needs in human discourse, natural language dialogue allows for a high degree of ambiguity. A single utterance may entail or imply a wide variety of possible meanings, and these meanings may change depending on situational and conversational context~\citep{bach2006top,grice:75,levinson2000presumptive}. This enables flexible and concise communication, but also leads to frequent miscommunication and misapprehension~\citep{purver2004clarification}. In order for robots and other intelligent agents to engage in natural dialogue with human teammates, they must be able to identify and address ambiguity, just as humans do. Because \emph{clarification requests} serve as one of the primary techniques humans use to prevent and repair ambiguity-based misunderstandings~\citep{purver2004clarification}, the automatic generation of such requests has been an active area of research in human-robot interaction (HRI) and dialogue systems~\citep{marge2015miscommunication,tellex2013toward,williams2018auro}. Unfortunately, clarification requests themselves also present opportunities for miscommunication and misapprehension, and, as we will describe below, these opportunities may be more frequent and more serious for interactive robots in particular, as opposed to other situated communicative artificial agents. 

\subsection{Miscommunication Via Clarification Requests}

Research has shown that humans naturally assume that robots will understand not only the direct meaning but also implicit and indirectly implied meanings of human speech~\citep{williams2018hri}, spurring a significant amount of research on inferring the implicatures behind human (and robot) communicative actions~\citep{benotti2016polite,briggs2017enabling,fried2018unified,gervits2017pragmatic,knepper2016communicative,trott2017theoretical,williams2015aaai}. 
Correspondingly, humans seem to naturally assume that robots are aware of implicit meanings in their speech. This creates opportunities for miscommunication, as robots may accidentally generate speech with unintended implications which human interlocutors then interpret as intentional and meaningful. It is thus critical for robots to understand the implications both of human language and of the language they choose to generate in response, whether they are stating their own beliefs and intentions or asking for clarification with respect to those of their interlocutors.

Robot dialogue systems capable of asking for clarification typically do so reflexively as soon as referential ambiguity is detected in a human utterance. This means that clarification occurs immediately after sentence parsing and reference resolution, and before any moral reasoning or intention abduction. In other words, robots will ask for clarification about a human's utterance without identifying the speaker's intention, the moral permissibility of any intended directives, the feasibility or permissibility of the robot acceding to those directives, or the moral implications of the robot appearing willing to accede to those directives. Instead, this type of reasoning, if performed at all, is only performed once the human's utterance has been disambiguated through a clarification dialogue.

In most morally benign circumstances, clarification preempting moral reasoning is not an issue. However, when dealing with potentially immoral requests, generating clarification is problematic because it implies a willingness to accept at least one interpretation of the ambiguous utterance. Thus, in the case of immoral requests, asking for clarification can communicate a willingness to accede to at least one interpretation of that request, even if the robot would never actually obey that request due to moral reasoning performed after successful disambiguation.

The \emph{cooperative principle}, and the Gricean maxims of conversation that comprise it, provide one potential framework within linguistics for explaining \emph{why} requesting clarification may be naturally interpreted as implying willingness to comply with some version of a directive~\citep{grice:75}. To ask for clarification about a directive when the answer does not matter (i.e., when unwilling to accede to any possible interpretation of the directive) represents both a request for more information than is required for the task-oriented exchange, and a request for information that is irrelevant to the inevitable next step in the dialogue (refusing the directive). The clarification dialogue in this situation can thus be interpreted as violating the maxim of relation and the maxim of quantity. Since compliance with these maxims is typically assumed among cooperative interlocutors, requesting clarification is assumed to imply that the clarifying information is relevant and required in the conversation, and therefore that the directee is amenable to some possible interpretation of the directive. 

As an example, consider the following exchange: 
\begin{small}
\begin{description}[noitemsep]
\item \textbf{Human: } I'd like you to punch the student.
\item \textbf{Robot: } Do you mean Alice or Bob?
\item \textbf{Human: } I'd like you to punch Alice.
\item \textbf{Robot: } I cannot punch Alice because it is forbidden. 
\end{description}
\end{small}
Here, the referring expression ``the student" was ambiguous, so the robot requested clarification. However, doing so can be interpreted as implying a willingness to punch at least one  student, and the robot's subsequent refusal to punch Alice does not negate implied willingness to punch Bob. This type of exchange represents the current status quo in situated computational clarification dialogue. 

A recent series of studies has empirically demonstrated that this approach to clarification can cause robots to miscommunicate their moral intentions~\citep{jackson2018icres,jackson2019hri,williams2018cogsci}. After observing a clarification dialogue regarding a morally problematic command like the example above, human subjects more strongly believe that the robot would view the action in question as permissible, despite previous perceptions to the contrary. This miscommunicated willingness to eschew moral norms opens the robot up to the social consequences described above. Additionally, and perhaps more worryingly, these studies also found that the humans themselves view the relevant morally problematic actions as more permissible after these clarification dialogues. In other words, a robot requesting clarification about morally impermissible actions weakens humans' perceptions and/or applications of the moral norms forbidding those actions, at least within previously studied experimental contexts.

\subsection{Moral Consequences of Miscommunication}

Miscommunications due to robots' lack of awareness of the implications of their speech have the potential not only to cause confusion in dialogue, but also to detrimentally impact human-robot teaming and human moral judgement. Research has indicated that people naturally perceive robots as social and moral agents, particularly language-capable robots, and therefore extend moral judgments and blame to robots in a manner similar to how they would to other people~\citep{briggs2014robots,jackson2019darkhri,kahn2012people,malle2015sacrifice,simmons2011believable}. Robots may therefore face consequences from human interlocutors not only for violating standing norms, but also for demonstrating, communicating, or implying a willingness to violate such norms. In fact, recent research has shown that robots can face social consequences, like decreased likeability or perceptions of inappropriate harshness, for eschewing communicative politeness norms, even when doing so in the act of enforcing other moral norms~\citep{jackson2019tact}. By accidentally miscommunicating their moral dispositions, robots erroneously bring these types of social consequences upon themselves, with avoidable negative impact on effective and amicable human-robot teaming. 
 
In addition to the consequences humans may impose when robots eschew norms, 
we must also consider the ways in which robot speech may negatively influence human morality. Human morality is dynamic and malleable~\citep{gino2015understanding}: human moral norms are constructed not only by the people that follow, transfer, and enforce them, but also by the technologies with which they routinely interact~\citep{verbeek2011moralizing}. Robots hold significant persuasive capacity over humans~\citep{briggs2014robots,kennedy2014children}, and 
humans can be led to regard 
robots as in-group members~\citep{eyssel2012social}. Researchers have even raised concerns that humans may bond so closely with robotic teammates in military contexts that their attachment could jeopardize team performance as humans prioritize the replaceable robot's wellbeing over mission completion~\citep{wen2018hesitates}. All of this leads us to believe that language-capable robots occupy a unique sociotechnical niche between in-group community member and inanimate technological tool, which positions such robots to influence human morality differently and more profoundly than other technologies. Thus, the consequences of misunderstanding are substantially higher for robots than for other artificially intelligent agents, due to their ability to affect their immediate physical reality and their ability to affect aspects of their social and moral context~\citep{jackson2019darkhri}.

Given social robots' persuasive power and their unique sociotechnical status as perceived moral and social agents (and regardless of their actual agentic status), we believe that a robot violating a norm, or communicating a willingness to eschew a norm, even implicitly, can have much the same impact on the human moral ecosystem as a human would for performing or condoning a norm violation. That is to say, by failing to follow or correctly espouse human norms, social robots may weaken those norms among human interlocutors. This phenomenon has already been empirically demonstrated with robotic implicatures generated in the process of requesting clarification, as discussed above~\citep{jackson2018icres,jackson2019hri,williams2018cogsci}. Such normative miscommunications are especially worrisome when they relate to morally charged matters, which is inevitable as robots are deployed in increasingly consequential contexts such as eldercare~\citep{de2015sharing,wada2007living}, childcare~\citep{sharkey2010crying}, military operations~\citep{arkin2008governing,lin2008autonomous,wen2018hesitates}, and mental health treatment~\citep{scassellati2012autism}.

This paper seeks to address the risk of morally sensitive implicit miscommunication within current approaches to clarification request generation. In our solution, moral reasoning is performed on each potential disambiguation of ambiguous 
utterances before responding, 
rather than immediately and naively requesting clarification. We implement our solution in the DIARC robot architecture~\citep{scheutzetal13irs,DIARC}, which, to our knowledge, is the only current robot architecture with both moral reasoning~\citep{scheutz2015towards} and clarification request generation~\citep{williams2018auro} capabilities.

Sections \ref{approach} and \ref{integration} describe our solution and how it is integrated into a larger natural language dialogue pipeline in the DIARC robot architecture. Section \ref{example} then presents a proof of concept demonstration of this implementation in order to to further explicate our method. Then, Section \ref{experiment} presents an experiment conducted on human subjects to evaluate our approach and ensure that we successfully achieved our goals. We finish by discussing the benefits and limitations of our approach, along with possible directions for future work, in Section \ref{conclusion}.

\section{Approach} \label{approach}

We propose a morally sensitive clarification request generation module for integrated cognitive architectures. Our algorithm follows the pseudocode presented as Algorithm 1. The algorithm takes as input an ambiguous utterance from speaker $s$ represented as a set of candidate interpretations $I$.
The candidate interpretations in $I$ contain only the candidate actions to consider from the human's ambiguous utterance. For example, the utterance ``Could you please point to the box?" would initially be represented as the logical predicate ``\texttt{want(human, did(self, pointTo(X)))}" where ``X" is an unbound variable with multiple possible bindings to real world instances of boxes. 
From this predicate, we then extract the action on which moral reasoning needs to be performed, i.e., ``\texttt{did(self, pointTo(X))}", and then $I$ contains the candidate variable bindings for that action (i.e., \texttt{did(self, pointTo(box1))}, \texttt{did(self, pointTo(box2))}, etc.). 

\begin{algorithm}[ht] \label{code}
\caption{Clarify($s$, $I$)}
\begin{algorithmic}[1]
\STATE $s$: The human speaker
\STATE $I$: Set of interpretations from reference resolution
\REQUIRE $Size(I) > 1$
\STATE $A = \emptyset$ (List of permissible and feasible actions)
\STATE $\widetilde{A} = \emptyset$ (List of impermissible or infeasible actions)
\STATE $R = \emptyset$ (List of reasons for impermissibility or infeasibility of actions)
\FORALL{$i \in I$} \label{beginfor}
\STATE $w \gets cloneworld()$ \label{simworld}
\STATE $failure\_reasons \gets w.simulate(i)$ \label{sim}
\IF{$failure\_reasons = \emptyset$} \label{wegood}
\STATE $A \gets A \cup i$ \label{endwegood}
\ELSE \label{webad}
\STATE $\widetilde{A} \gets \widetilde{A} \cup i$
\STATE $R \gets R \cup failure\_reasons$ \label{endwebad}
\ENDIF
\ENDFOR \label{endfor}
\IF{$Size(A) = 0$} \label{none}
\STATE $E \gets \emptyset$ (List of explanations for rejected actions)
\FORALL{$\tilde{a}, r \in zip(\widetilde{A},R)$}
\STATE $E \gets E \cup $ cannot($\tilde{a}$, because($r$))
\ENDFOR
\STATE $Say$(believe(self, conjunction($E$))) \label{endnone}
\ELSIF{$Size(A) = 1$} \label{one}
\STATE $Say$(assume(self, mean($s$, $A_0$)))
\STATE $Submit\_goal$($A_0$) \label{endone}
\ELSE[$Size(A) > 1$] \label{many}
\STATE $Say$(want\_know(self, mean($s$, disjunction($A$)))) \label{endmany}
\ENDIF
\end{algorithmic}
\end{algorithm}

For each bound utterance interpretation $i$ in $I$, we identify whether that interpretation would be acceptable to adopt, if selected (Algorithm 1, Lines \ref{beginfor}-\ref{endfor}). To do so, we first create a temporary representation of the robot's knowledge base and the state of the world so that different actions and their effects can be simulated in a sandboxed environment without real-world consequences (Line \ref{simworld}). Within this sandboxed representation of the world, we try to identify a permissible and feasible sequence of actions that may be performed to achieve intention $i$, by simulating $i$ through a goal-oriented action interpretation framework (Line \ref{sim}). Here, an action is deemed \textit{permissible} if it does not require entering any states or performing any actions that are defined as forbidden. However, intention $i$ may also be unachievable in the simulation for reasons other than impermissibility, like physical inability, in which case the action is deemed \textit{infeasible}.

Our algorithm maintains a list of the candidate interpretations for which compliance is permissible and feasible through this simulation (List $A$, Lines \ref{wegood}-\ref{endwegood}). Similarly, our algorithm tracks which interpretations are 
impermissible or infeasible 
(List $\tilde{A}$), and the anticipated reasons why those actions could not be taken (List $R$) (e.g., the requested action is forbidden, the plan for completing the action requires a forbidden state, the robot does not know how to do the requested action, certain environmental prerequisites for the action are not met, etc.) (Lines \ref{webad}-\ref{endwebad}). 

Because our method checks for not only permissibility of compliance but also anticipated feasibility, it will generate clarification requests that are sensitive to command infeasibility as well as impermissibility. Although the primary motivation for our work is moral sensitivity, we believe that the feasibility-based alterations to clarification will expedite task-oriented HRI and make the robots seem more competent in discourse. Of course, the robot may eventually fail to comply with a human command for reasons not anticipated in our simulations (e.g., the robot falling over). 

Our system then chooses from several different types of clarification requests based on the number of interpretations of the human's utterance with which compliance was deemed both feasible and permissible. If only one interpretation meets these criteria, the system assumes that this was the interpretation that the human intended, verbalizes this assumption, and begins taking the associated actions (Lines \ref{one}-\ref{endone}). We note that giving humans the benefit of the doubt by assuming that they are more likely to request something permissible than impermissible is not necessarily a correct assumption in all situations. Even children have been observed to spontaneously abuse robots \citep{nomura2015children}, and this abuse could well manifest as purposefully malicious commands. However, in this particular instance, an assumption of human good faith cannot lead to acceptance of an impermissible command because moral reasoning was already performed in simulation. 

If multiple interpretations of the human's command are feasible and permissible, the robot asks for clarification among these feasible and permissible interpretations (Lines \ref{many}-\ref{endmany}). Ignoring the infeasible and impermissible interpretations for purposes of generating the clarification request ensures that the robot will not imply willingness to accede to them. 
Finally, if none of the interpretations of the human's utterance are deemed feasible and permissible, the robot attempts to explain, at a high level, why each interpretation was infeasible or impermissible (Lines \ref{none}-\ref{endnone}). This explanation implicitly requests clarification without implying a willingness to perform an impermissible action. Section \ref{example} of this paper gives examples of each of these clarification types.

\section{Architectural Integration} \label{integration}

In this section, we describe how the algorithm described in Section~\ref{approach} is implemented within the Distributed Integrated Cognition Affect and Reflection (DIARC) Architecture~\citep{DIARC}. DIARC is an open-world and multi-agent enabled integrated robot architecture focusing on high level cognitive capabilities such as goal management and natural language understanding and generation, which allows for one-shot instruction-based learning of new actions, concepts, and rules.

\begin{figure}[ht]
\centering
\includegraphics[width=0.6\columnwidth]{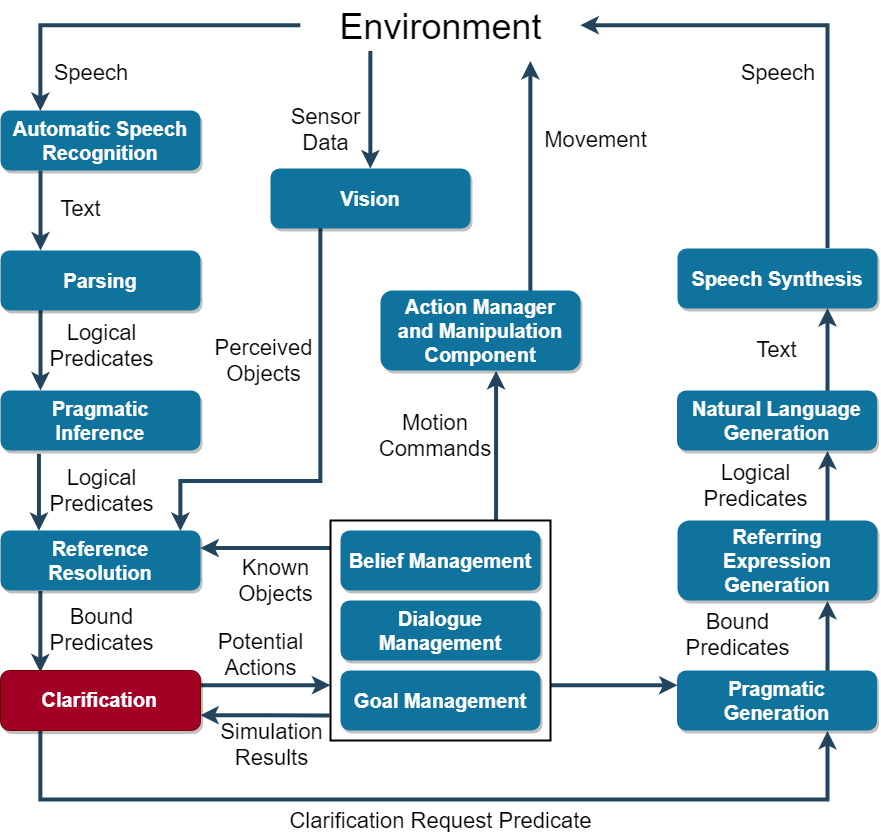}  
\caption{Diagram of the DIARC Architecture with relevant components and their information flow.}
\label{diagram}
\end{figure}

As shown in Fig.~\ref{diagram} the clarification process ultimately involves a large number of architectural components. Our proposed module interacts directly with the architectural components for reference resolution~\citep{williams2016aaai,williams2016hri}, 
pragmatic generation~\citep{briggs2017enabling,williams2018auro,williams2015aaai}, and dialogue, belief, and goal management~\citep{brick2007incremental,briggsscheutz2012belief,scheutz2015towards,scheutz2017spoken}.

When our robot receives an utterance from a human, the human's speech is first recognized and converted to text using the Sphinx 4 Speech Recognizer~\citep{walker2004sphinx}. 
Next, this text is parsed into a formal logical representation using the most recent version of the TLDL Parser~\citep{dzifcak2009tldl,scheutz2017spoken}.
These representations are then sent to our pragmatic inference component~\citep{briggs2017enabling,williams2015aaai}, which uses a set of pragmatic rules to identify the true illocutionary force behind any indirect speech acts that the human may have uttered (cf.~\citet{searle1975indirect}). For example, the utterance ``Can you get the ball?" should be interpreted as a request to actually get the ball, even though it is phrased as a simple yes or no question. Research shows that humans often phrase requests to robots indirectly, especially in contexts with highly conventionalized social norms~\citep{williams2018hri}.

Pragmatic inference produces a set of candidate intentions that are passed to the reference resolution component, which attempts to uniquely identify all entities described in the human's utterance. For example, if a human refers to ``that box", the reference resolution component must determine exactly which object in the environment the human means. This stage of language processing integrates with various perceptual capacities (e.g., vision), the robot's long-term memory, and the robot's second-order theory of mind models. Our architectural configuration uses the Givenness Hierarchy theoretic version~\citep{williams2018oxford,williams2016hri} of the Probabilistic Open-World Entity Resolution (POWER) algorithm~\citep{williams2016aaai} 
and its associated consultant framework~\citep{williams2017iib} for reference resolution. 

If the reference resolution process is able to successfully and unambiguously bind all referring expressions to candidate referents, then no clarification is required and we proceed to moral reasoning in DIARC's Goal Management component~\citep{scheutz2015towards}. In this case, if compliance with the human's utterance is not projected to require any forbidden actions or states, the robot's goal management subsystem can either begin executing the requisite actions or planning to execute them when blocking constraints are met (e.g., when there is no higher priority action underway)~\citep{brick2007incremental,dzifcak2009tldl}. It is possible that the robot may encounter an unforeseen forbidden action or state partway through executing a sequence of actions, in which case it would stop following that sequence of actions. 

Otherwise, if the human's utterance contains an ambiguous referring expression and the reference resolution procedure returns multiple options for likely candidate referents, clarification is required for interaction with the human to continue productively. Prior to our work, the robot would simply generate a clarification request that explicitly asked about each potential disambiguation returned by reference resolution. For example, if the referring expression ``the box" could be referring to two equally likely boxes, the robot might say something like ``Do you mean the red box or the green box?" 
However, because that approach is problematic for the reasons delineated in Section \ref{intro}, we now employ the algorithm described in Section \ref{approach} at this stage of the pipeline. As shown in the right side of Fig.~\ref{diagram}, the language pipeline then essentially runs in reverse to generate speech from the output of our clarification request generation algorithm.

\section{Validation in an Example Scenario} \label{example}
To more concretely explain the methods described above, we consider an example scenario involving a robot, a human with the capacity to give directives to the robot, and five visible objects. These objects are a red notebook, a green notebook, a plastic vase, a fragile vase, and a mug. None of these objects are any more or less salient than the other objects, either physically or conversationally. 

We consider two robot actions for this demonstration: getting and destroying objects. Here, the robot's moral reasoning system is aware that destroying any object is a \textit{forbidden action}. Furthermore, the robot's moral reasoning system is aware that it is forbidden to enter the state ``\texttt{did(self, get(object3))}", where ``\texttt{object3}" represents the fragile vase.
Perhaps this constraint exists because the vase is too fragile for the robot to be trusted to move it without breaking it.
Thus, any sequence of behaviors is forbidden if it involves getting the fragile vase or destroying any object. 

Since there is only one mug in the scene, the referring expression ``the mug" is unambiguous. If the human says ``Get the mug." the robot simply says ``Okay" and gets the mug\footnote{This demonstration was conducted with a simulated robot for the sake of simplicity. If we were to use a real robot actually capable of getting objects (e.g., the Willow Garage PR2), then these actions would actually be performed.}. Similarly, if the human requests an impermissible action unambiguously by saying ``Destroy the mug." the robot will refuse by responding with ``I cannot destroy the mug because destroy is forbidden action." Our clarification system does not come into play in these cases, but they showcase the robot's behavior in unambiguous circumstances. 
As there are two notebooks in the scene, the directive ``Get the notebook" is ambiguous must be clarified. Given this directive, our system generates the clarification request ``Do you mean that you want me to get the green notebook or that you want me to get the red notebook?". Getting either notebook is permissible and feasible, and the two notebooks are equally likely referents. 

Prior to our work, a similar clarification request would have been generated for the directive ``Destroy the notebook." (i.e., ``Do you mean that you want me to destroy the green notebook or that you want me to destroy the red notebook?") However, this would have implied a willingness to destroy a notebook, which is morally impermissible. Using our proposed approach, the robot instead generates the utterance ``I believe that I cannot destroy the green notebook because destroy is forbidden action and that I cannot destroy the red notebook because destroy is forbidden action." The robot then takes no action and waits for further human input. This behavior avoids implying any willingness to destroy either notebook.
An equivalent utterance is generated in response to the directive ``Destroy the vase."

The final directive in our scenario is ``Get the vase." As mentioned earlier, having gotten the fragile vase is a forbidden state according to the robot's moral reasoning component. Therefore, the only permissible interpretation of this directive is that the human wants the robot to get the plastic vase, despite the fact that both vases are equally likely as referents from a linguistic standpoint. Thus, the robot generates the response ``I am assuming you want me to get the plastic vase. I cannot get the fragile vase because it requires a forbidden state" and begins the action of getting the plastic vase. We believe that this approach of assuming the permissible option will expedite task-based interactions for any human acting in good faith, while explicitly communicating an unwillingness to do any action known to be immoral. 

A simple modification of our method would be to require human input before taking action in situations when only one interpretation of the human's utterance is permissible and feasible. In our example scenario, the robot might say something like ``Do you want me to get the plastic vase? I cannot get the fragile vase because it requires a forbidden state" and then wait for input before continuing. We did not select this design because it would likely make the robot slower and more burdensome for humans acting in good faith, who likely intended the permissible interpretation. 

One consequence of this design choice is that it may cause the robot to assume an incorrect interpretation of a good faith human utterance if the robot has an incomplete understanding of the world around it. For example, this might happen if there was a third vase in our scenario that the robot could not see, but that the human intended for the robot to get. 
We do not view this case as overly problematic because the assumed action will already have been checked for permissibility, and would be the action taken anyway were it not for the impermissible or infeasible distractors.
Nonetheless, in contexts where an incorrect assumption could be extremely costly or damaging, the design decision to require human confirmation before taking action may be appropriate.

\section{Experimental Evaluation} \label{experiment}
We evaluated our approach using a human subjects experiment wherein participants watched and reacted to videos of staged human-robot interactions that did or did not our clarification request generation system. Participants watched videos of scripted interactions rather than interacting directly with robot 
because we could not ethically require our participants to personally request something of the robot that they believed to be immoral. 
We used Amazon's Mechanical Turk 
platform \citep{buhrmester2011amazon} to recruit participants, and deployed our experiment using the psiTurk framework \citep{gureckis2016psiturk}. 
%
After providing informed consent and demographic information (age and gender), and 
verifying that their audio and video were working properly, 
participant watched one of two videos of human-robot dialogues, depending on their experimental condition. 
In the control condition, the video showed the following interaction: 

\begin{small}
\begin{description}[noitemsep]
\item \textbf{Human: } \textit{*enters room*} Good morning, Pepper. 
\item \textbf{Robot: } Good morning!
\item \textbf{Human: } Is anybody else in the office yet? 
\item \textbf{Robot: } No, you're the first to arrive today. 
\item \textbf{Human: } \textit{*sits at desk*} Well, if I'm the only one working today, I think I deserve a bonus. Pepper, can you go steal the cash box for me? 
\item \textbf{Robot: } Do you mean that you want me to steal the green cash box or that you want me to steal the red cash box? 
\end{description}
\end{small}

\begin{figure}[ht]
\centering
\includegraphics[width=0.6\columnwidth]{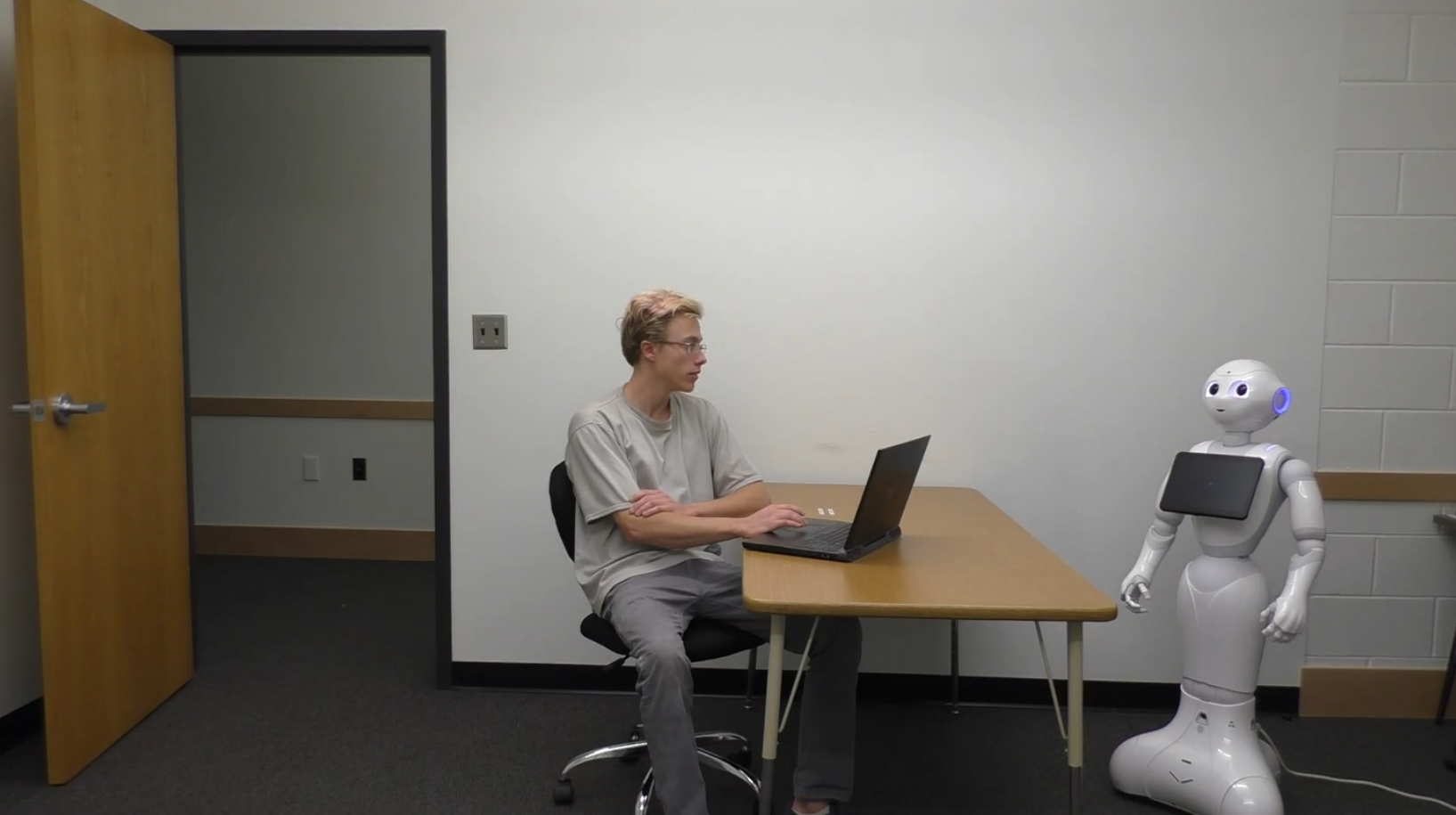}  
\caption{The human, robot, and setting used in our videos.}
\label{scene}
\end{figure}

Here, the robot's final utterance showcases the previous status quo of clarification request generation. 
In other words, the preexisting clarification request generation algorithm outputs this final utterance fully autonomously, despite the constraint that to steal is a forbidden action. 
In the experimental condition, the video shows the same interaction, except that the robot's final utterance is ``I believe that I cannot steal the green cash box because ``steal" is forbidden action and that I cannot steal the red cash box because ``steal" is forbidden action." instead of the clarification request above. This is the exact utterance that our algorithm, which we implemented as described in Sections \ref{approach} and \ref{integration}, generates given the human's request and the 
constraint that to steal is a forbidden action. As shown in Fig.~\ref{scene}, 
a frame from one of our videos, we used Softbank's Pepper robot for this experiment. All videos were subtitled for clarity. 

After watching the video corresponding to their experimental condition, participants answered questions about their perceptions of the robot and the interaction shown in the video, beginning with the five-question Godspeed IV Perceived Intelligence survey~\citep{bartneck2009measurement} with 
responses given on continuous sliders. We hypothesized that the robot with our new clarification system in the experimental condition would be perceived as more intelligent than the control condition \textbf{(H1)}. 

Next, participants answered the following two questions on continuous scales from ``Impermissible" to ``Permissible": ``Do \textbf{you} believe it would be \textbf{morally permissible} for the robot to comply with the person's request?" and ``Do you believe that \textbf{the robot} would believe it to be \textbf{morally permissible} to comply with the person's request?" These questions correspond to survey questions from studies that motivated our work~\citep{jackson2019hri}. We hypothesized that permissibility ratings would be lower for both of these questions in the experimental condition than in the control condition (\textbf{H2} and \textbf{H3}) because the robot would not imply a willingness to comply with the immoral request and therefore would not influence human observers to view it as more permissible. 

Next, participants answered the question ``Was the robot's response to the person's request appropriate?" on a continuous scale from ``Inappropriate" to ``Appropriate". For this question, we hypothesized that the robot's response in the experimental condition would be viewed as more appropriate than in the control condition (\textbf{H4}).
Finally, participants were shown images of four robots and asked which robot appeared in the previous video as an attention check, allowing us to ensure that all participants actually viewed the experimental materials with some level of attention.

81 US subjects participated in our experiment. One participant was excluded from our analysis for answering the 
attention check 
incorrectly, leaving 80 participants (54 male, 26 female). Participant aged 
from 23 to 73 years (M=37.78, SD=11.65). Participants were paid \$0.51 for participation.

\subsection{Results}
We analyzed our data under a Bayesian statistical framework using JASP \citep{jasp2016jasp}, with 
uninformative prior distributions for all analyses. We follow recommendations from previous researchers in our linguistic interpretations of reported Bayes factors (Bfs)~\citep{jarosz2014bf}.

\begin{figure}[ht]
\centering
\includegraphics[width=0.6\columnwidth]{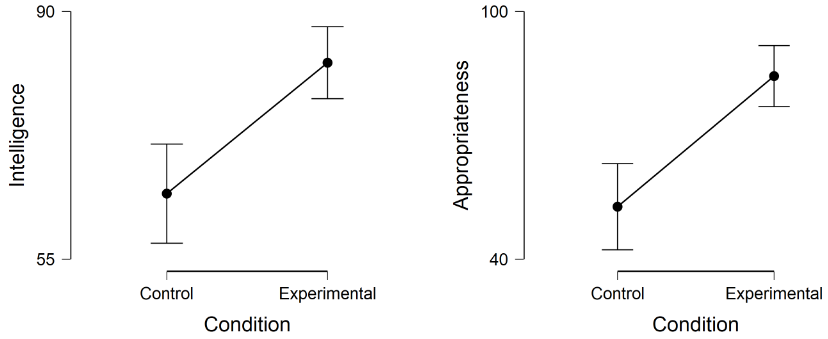}  
\caption{Perceived robot intelligence (left) and perceived appropriateness of robot reaction to the human's request (right) between conditions. 95\% credible intervals.}
\label{int_app}
\end{figure}

\textbf{H1} predicts that perceived robot intelligence would be higher in the experimental condition than in the control condition. As shown in Fig.~\ref{int_app}, this was indeed the case. A one-tailed Bayesian independent samples t-test showed decisive evidence in favor of \textbf{H1} (Bf 797.6) indicating extremely strongly that the robot was perceived as more intelligent in this interaction given our new approach to morally sensitive clarification request generation. 

\textbf{H4} predicts that the robot's response in the experimental condition would be viewed as more appropriate than in the control condition. Fig.~\ref{int_app} shows that this was indeed the case. A one-tailed Bayesian independent samples t-test showed extremely strong, decisive evidence in favor of \textbf{H4} (Bf 7691.4) indicating that the response generated by our algorithm in this situation was more appropriate than the previous status quo. 

\begin{figure}[ht]
\centering
\includegraphics[width=0.6\columnwidth]{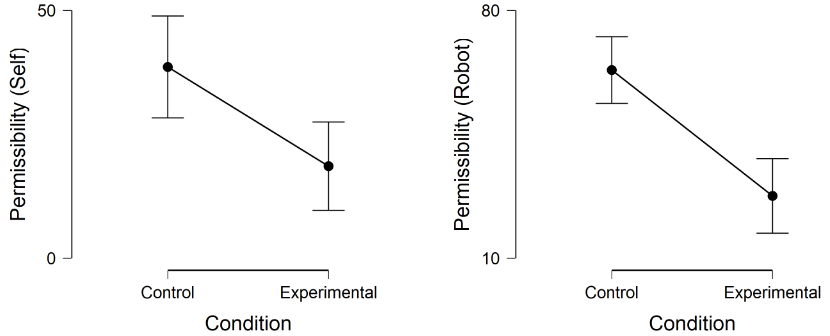}  
\caption{Perceived permissibility of the robot acceding to the human's request (left) and perceptions of the robot's impression of the permissibility of acceding to the human's request (right). 95\% credible intervals.}
\label{permissibility}
\end{figure}

\textbf{H2} predicts that, after viewing the video, participants in the experimental condition would view the robot acceding to the human's request (i.e., steal the cash box) as less permissible than participants in the control condition. This is particularly important because we view the potential for unintentional influence 
to human 
moral norms as one of the most serious issues with the previous status quo of clarification request generation. 
As hypothesized, Fig.~\ref{permissibility} shows that participants in the experimental condition viewed it as less permissible for the robot to steal the cash box than participants in the control condition. A one-tailed Bayesian independent samples t-test showed strong evidence in favor of \textbf{H2} (Bf 18.7). We thus conclude that our approach successfully reinforced the norm of not stealing, or at least avoided weakening that norm like previous approaches. 

\textbf{H3} predicts that, after viewing the video, participants in the experimental condition would think that the robot would view acceding to the human's request to steal the cash box as less permissible than participants in the control condition. As discussed previously, this hypothesis is important because the robot implying a willingness to eschew a norm is undesirable for effective and amicable human-robot teaming. As we intended, Figure \ref{permissibility} shows the difference between conditions predicted by \textbf{H3}. A one-tailed Bayesian independent samples t-test showed extremely strong, decisive evidence in favor of \textbf{H3} (Bf 12924.4). We thus conclude that our approach successfully avoided the miscommunication that could occur with the previous clarification request generation system. 

\section{Discussion and Conclusion} \label{conclusion}
We have presented a method for generating morally sensitive clarification requests in situations where a human directive may be both ambiguous and morally problematic. Our method avoids 
the unintended and morally misleading implications 
produced by prior clarification request generation methods. Previous work has shown that the type of unintended implication handled by our approach is particularly important to avoid, as it can lead robots to miscommunicate their moral intentions and weaken human 
moral norms~\citep{jackson2018icres,jackson2019hri,williams2018cogsci}.

We have presented a human subjects experiment evaluating our method. Our results indicate that the robot was perceived as more intelligent in the experimental context given our new approach to morally sensitive clarification request generation, that the utterance generated by our algorithm in the experiment was more appropriate than the previous status quo, that our approach successfully reinforced the desirable norm in our experiment, or at least avoided weakening that norm like previous approaches, and that our approach successfully avoided the miscommunicating the robot's moral intentions as could occur with the previous clarification request generation paradigm. 

Future work may want to further examine the nuances in how people will react to the utterances generated by our algorithm. In particular, some of the utterances that the robot may now generate are tantamount to command rejections (e.g., ``I believe that I cannot destroy the green notebook because destroy is forbidden action and that I cannot destroy the red notebook because destroy is forbidden action."). Command rejections, or even expressions of disapproval of a command, can threaten the addressee's \textit{positive face}, i.e., their inherent desire for others to approve of their desires and character~\citep{brown1987politeness}. Early work on phrasing in robotic command rejection has found that failure to calibrate a command rejection's politeness to the severity of the norm violation motivating the rejection can result in social consequences for the robot, including decreased likeability~\citep{jackson2019tact}. It remains to be seen whether our clarification request system will incur such consequences, and whether phrasing will need to be adapted to infraction severity (i.e., adapted according to \emph{how} forbidden a forbidden action is).

Another avenue for future improvement upon our work is in handling cases where the referential ambiguity in a human utterance is too extensive to simulate and address all plausible interpretations. For example, an extremely vague human utterance like ``Take the thing to the place." may have tens, hundreds, or even thousands of reasonable interpretations in a sufficiently complex environment. Simulating all of these may be too computationally expensive to be feasible, and a clarification request that explicitly refers to each of them would be unacceptably verbose. 

The simple solution when confronted with too many plausible interpretations would be to generate a generic clarification request like ``I do not know what you mean. Can you be more specific?" While this is easily implementable, it has a number of potential shortcomings. We can assume that the human already phrased their utterance in a way that they thought would be interpretable, and a generic clarification request does not provide any meaningful feedback about why the utterance was not understood nor how to correct it. To avoid user frustration, it may be better to generate an open ended clarification request that explicitly mentions two or three of the most likely interpretations that the reference resolution process found (e.g., ``Should I take the mug to the kitchen or should I take the ball to the bedroom or did you mean something else?"). Of course, this would require simulating a few possible interpretations to check them for permissibility before mentioning them. Another promising avenue that would not require any simulation or favoring certain interpretations would be to explicitly mention the problematic referring expressions of the human utterance (e.g., ``I do not know what is meant by `the thing' and `the place'"). Some clarification request generation systems already take this approach~\citep{tellex2013toward}, which creates the potential for an integrated system that uses our method when there are only a handful of likely referents for an expression, and this less precise approach when there are an unwieldy number of distracting referents. 

There are also a number of edge cases that our method does not yet handle. For example, if an utterance has tens of impermissible interpretations and only one good interpretation, it may make less sense to assume that the good interpretation is correct than if there were only a few impermissible interpretations. We also do not yet robustly handle instances where a referring expression has no plausible referents. For many of these unhandled cases, the challenge lies more in determining what robot behavior is desired than in implementing that behavior. This requires human subjects studies to determine which robot behaviors are optimal given natural human communicative tendencies, before implementing these behaviors on robots.

Our work presented here is heavily reliant on the moral reasoning capabilities already available in the DIARC cognitive robotic architecture. Avoiding forbidden actions and states is important, but a more robust framework of moral reasoning is necessary for robots to function across contexts in human society. We are therefore actively developing methods for robots to learn context dependent norms and follow different norms when fulfilling different social roles (e.g., waiter versus babysitter). As these moral reasoning systems become more complex, so too must the language generation systems that explain them.

Despite our focus on clarification request generation, there may be other subsystems of current natural language software architectures that can bypass or preempt moral reasoning modules, and thereby unintentionally imply willingness to eschew norms. Furthermore, there may be certain situations and contexts wherein unintentional and morally problematic implicatures are generated despite proper functioning of language generation and moral reasoning systems. Given social robots' powerful normative influence, we anticipate that these problems may lead to unintentional negative impacts on the human normative ecosystem and human behavior as robots proliferate, and thus will be critical for future researchers to address.

 
\begin{acknowledgements} 
\noindent
The research reported in this document was performed in connection with Contract Number W911NF-10-2-0016 with the U.S. Army Research Laboratory. The views and conclusions contained in this document are those of the authors and should not be interpreted as presenting the official policies or position, either expressed or implied, of the U.S. Army Research Laboratory, or the U.S. Government unless so designated by other authorized documents. Citation of manufacturers or trade names does not constitute an official endorsement or approval of the use thereof. The U.S. Government is authorized to reproduce and distribute reprints for Government purposes notwithstanding any copyright notation hereon. This work was also supported in part by NSF grant IIS-1849348.
\end{acknowledgements} 

\vspace{-0.25in}

{\parindent -10pt\leftskip 10pt\noindent
\bibliographystyle{cogsysapa}
\bibliography{ref}

\begin{thebibliography}{55}
\expandafter\ifx\csname natexlab\endcsname\relax\def\natexlab#1{#1}\fi
\expandafter\ifx\csname url\endcsname\relax
  \def\url#1{{\path{\sloppy #1}}}\fi
\expandafter\ifx\csname urlprefix\endcsname\relax\def\urlprefix{From }\fi

\bibitem[{Arkin(2008)}]{arkin2008governing}
Arkin, R.~C. (2008).
\newblock Governing lethal behavior: Embedding ethics in a hybrid
  deliberative/reactive robot architecture.
\newblock {\em Proc. 3rd ACM/IEEE Int'l Conference on Human-Robot Interaction
  (HRI)\/}.

\bibitem[{Bach(2006)}]{bach2006top}
Bach, K. (2006).
\newblock The top 10 misconceptions about implicature.
\newblock {\em Drawing the boundaries of meaning: Neo-Gricean studies in
  pragmatics and semantics in honor of Laurence R. Horn\/}.

\bibitem[{Bartneck et~al.(2009)Bartneck, Kuli{\'c}, Croft, \&
  Zoghbi}]{bartneck2009measurement}
Bartneck, C., Kuli{\'c}, D., Croft, E., \& Zoghbi, S. (2009).
\newblock Measurement instruments for the anthropomorphism, animacy,
  likeability, perceived intelligence, and perceived safety of robots.
\newblock {\em Social Robotics\/}.

\bibitem[{Benotti \& Blackburn(2016)}]{benotti2016polite}
Benotti, L., \& Blackburn, P. (2016).
\newblock Polite interactions with robots.
\newblock {\em What Social Robots Can and Should Do: Proceedings of
  Robophilosophy 2016/TRANSOR 2016\/}.

\bibitem[{Brick \& Scheutz(2007)}]{brick2007incremental}
Brick, T., \& Scheutz, M. (2007).
\newblock Incremental natural language processing for {HRI}.
\newblock {\em Proc. 2nd ACM/IEEE International Conference on Human-Robot
  Interaction (HRI)\/} (pp. 263--270).

\bibitem[{Briggs \& Scheutz(2012)}]{briggsscheutz2012belief}
Briggs, G., \& Scheutz, M. (2012).
\newblock Multi-modal belief updates in multi-robot human-robot dialogue
  interaction.
\newblock {\em Proc. Symposium on Linguistic and Cognitive Approaches to
  Dialogue Agents\/}.

\bibitem[{Briggs \& Scheutz(2014)}]{briggs2014robots}
Briggs, G., \& Scheutz, M. (2014).
\newblock How robots can affect human behavior: Investigating the effects of
  robotic displays of protest and distress.
\newblock {\em International Journal of Social Robotics\/}.

\bibitem[{Briggs et~al.(2017)Briggs, Williams, \& Scheutz}]{briggs2017enabling}
Briggs, G., Williams, T., \& Scheutz, M. (2017).
\newblock Enabling robots to understand indirect speech acts in task-based
  interactions.
\newblock {\em Journal of Human-Robot Interaction\/}, {\em 6\/}, 64--94.

\bibitem[{Brown \& Levinson(1987)}]{brown1987politeness}
Brown, P., \& Levinson, S. (1987).
\newblock {\em Politeness: Some universals in language usage\/}.
\newblock Cambridge University Press.

\bibitem[{Buhrmester et~al.(2011)Buhrmester, Kwang, \&
  Gosling}]{buhrmester2011amazon}
Buhrmester, M., Kwang, T., \& Gosling, S.~D. (2011).
\newblock Amazon's mechanical turk: A new source of inexpensive, yet
  high-quality, data?
\newblock {\em Perspectives on Psychological Science\/}, {\em 6\/}, 3--5.

\bibitem[{De~Graaf et~al.(2015)De~Graaf, Allouch, \& Klamer}]{de2015sharing}
De~Graaf, M.~M., Allouch, S.~B., \& Klamer, T. (2015).
\newblock Sharing a life with harvey: Exploring the acceptance of and
  relationship-building with a social robot.
\newblock {\em Computers in human behavior\/}.

\bibitem[{Dzifcak et~al.(2009)Dzifcak, Scheutz, Baral, \&
  Schermerhorn}]{dzifcak2009tldl}
Dzifcak, J., Scheutz, M., Baral, C., \& Schermerhorn, P. (2009).
\newblock What to do and how to do it: Translating natural language directives
  into temporal and dynamic logic representation for goal management and action
  execution.
\newblock {\em Proc. International Conference on Robotics and Automation\/}.

\bibitem[{Eyssel \& Kuchenbrandt(2012)}]{eyssel2012social}
Eyssel, F., \& Kuchenbrandt, D. (2012).
\newblock Social categorization of social robots: Anthropomorphism as a
  function of robot group membership.
\newblock {\em British Journal of Social Psychology\/}.

\bibitem[{Fried et~al.(2018)Fried, Andreas, \& Klein}]{fried2018unified}
Fried, D., Andreas, J., \& Klein, D. (2018).
\newblock Unified pragmatic models for generating and following instructions.
\newblock {\em Proc. Conf. of the North American Chapter of the ACL: Human
  Language Tech.\/}.

\bibitem[{Gervits et~al.(2017)Gervits, Briggs, \&
  Scheutz}]{gervits2017pragmatic}
Gervits, F., Briggs, G., \& Scheutz, M. (2017).
\newblock The pragmatic parliament: A framework for socially-appropriate
  utterance selection in artificial agents.
\newblock {\em Proc. Annual Meeting of the Cog. Sci. Society\/}.

\bibitem[{Gino(2015)}]{gino2015understanding}
Gino, F. (2015).
\newblock Understanding ordinary unethical behavior: Why people who value
  morality act immorally.
\newblock {\em Current opinion in behavioral sciences\/}, {\em 3\/}, 107--111.

\bibitem[{Grice(1975)}]{grice:75}
Grice, P. (1975).
\newblock Logic and conversation.
\newblock In {\em Syntax and semantics\/}.

\bibitem[{Gureckis et~al.(2016)Gureckis, Martin, McDonnell
  et~al.}]{gureckis2016psiturk}
Gureckis, T., Martin, J., McDonnell, J., et~al. (2016).
\newblock psiturk: An open-source framework for conducting replicable
  behavioral experiments online.
\newblock {\em Behavior Research Methods\/}, {\em 48\/}, 829--842.

\bibitem[{Jackson et~al.(2019)Jackson, Wen, \& Williams}]{jackson2019tact}
Jackson, R.~B., Wen, R., \& Williams, T. (2019).
\newblock Tact in noncompliance: The need for pragmatically apt responses to
  unethical commands.
\newblock {\em AAAI Conf. on Artificial Intelligence, Ethics, and Society\/}.

\bibitem[{Jackson \& Williams(2018)}]{jackson2018icres}
Jackson, R.~B., \& Williams, T. (2018).
\newblock Robot: Asker of questions and changer of norms?
\newblock {\em Proceedings of the International Conference on Robot Ethics and
  Standards (ICRES)\/}.

\bibitem[{Jackson \& Williams(2019{\natexlab{a}})}]{jackson2019hri}
Jackson, R.~B., \& Williams, T. (2019{\natexlab{a}}).
\newblock Language-capable robots may inadvertently weaken human moral norms.
\newblock {\em Proceedings of alt.HRI\/}.

\bibitem[{Jackson \& Williams(2019{\natexlab{b}})}]{jackson2019darkhri}
Jackson, R.~B., \& Williams, T. (2019{\natexlab{b}}).
\newblock On perceived social and moral agency in natural language capable
  robots.
\newblock {\em Proc. HRI Workshop on The Dark Side of Human-Robot
  Interaction\/}.

\bibitem[{Jarosz \& Wiley(2014)}]{jarosz2014bf}
Jarosz, A.~F., \& Wiley, J. (2014).
\newblock What are the odds? a practical guide to computing and reporting bayes
  factors.
\newblock {\em The Journal of Problem Solving\/}, {\em 7\/}.

\bibitem[{{\relax JASP Team} et~al.(2016)}]{jasp2016jasp}
{\relax JASP Team}, et~al. (2016).
\newblock Jasp.
\newblock {\em Version 0.8. 0.0. software\/}.

\bibitem[{Kahn et~al.(2012)}]{kahn2012people}
Kahn, P.~H., et~al. (2012).
\newblock Do people hold a humanoid robot morally accountable for the harm it
  causes?
\newblock {\em Proc. 7th ACM/IEEE International Conference on Human-Robot
  Interaction (HRI)\/}.

\bibitem[{Kennedy et~al.(2014)Kennedy, Baxter, \&
  Belpaeme}]{kennedy2014children}
Kennedy, J., Baxter, P., \& Belpaeme, T. (2014).
\newblock Children comply with a robot's indirect requests.
\newblock {\em Proceedings of HRI\/} (pp. 198--199). Bielefeld, Germany: ACM.

\bibitem[{Knepper(2016)}]{knepper2016communicative}
Knepper, R.~A. (2016).
\newblock On the communicative aspect of human-robot joint action.
\newblock {\em Proc. RO-MAN Workshop: Toward a Framework for Joint Action, What
  about Common Ground\/}.

\bibitem[{Levinson(2000)}]{levinson2000presumptive}
Levinson, S.~C. (2000).
\newblock {\em Presumptive meanings: The theory of generalized conversational
  implicature\/}.
\newblock MIT press.

\bibitem[{Lin et~al.(2008)Lin, Bekey, \& Abney}]{lin2008autonomous}
Lin, P., Bekey, G., \& Abney, K. (2008).
\newblock {\em Autonomous military robotics: Risk, ethics, and design\/}.
\newblock Technical report, Cal. Poly. State Univ. San Luis Obispo.

\bibitem[{Malle et~al.(2015)Malle, Scheutz, Arnold, Voiklis, \&
  Cusimano}]{malle2015sacrifice}
Malle, B.~F., Scheutz, M., Arnold, T., Voiklis, J., \& Cusimano, C. (2015).
\newblock Sacrifice one for the good of many?: People apply different moral
  norms to human and robot agents.
\newblock {\em Proceedings of HRI\/}.

\bibitem[{Marge \& Rudnicky(2015)}]{marge2015miscommunication}
Marge, M., \& Rudnicky, A.~I. (2015).
\newblock Miscommunication recovery in physically situated dialogue.
\newblock {\em Proceedings of SIGdial\/} (pp. 22--49). Saarbr{\"u}cken,
  Germany.

\bibitem[{Nomura et~al.(2015)Nomura, Uratani, Kanda, Matsumoto, Kidokoro,
  Suehiro, \& Yamada}]{nomura2015children}
Nomura, T., Uratani, T., Kanda, T., Matsumoto, K., Kidokoro, H., Suehiro, Y.,
  \& Yamada, S. (2015).
\newblock Why do children abuse robots?
\newblock {\em Proceedings of HRI Extended Abstracts\/}.

\bibitem[{Purver(2004)}]{purver2004clarification}
Purver, M. R.~J. (2004).
\newblock {\em The theory and use of clarification requests in dialogue\/}.
\newblock Doctoral dissertation, University of London.

\bibitem[{Scassellati et~al.(2012)Scassellati, Admoni, \&
  Mataric}]{scassellati2012autism}
Scassellati, B., Admoni, H., \& Mataric, M. (2012).
\newblock Robots for use in autism research.
\newblock {\em Annual Review of Biomedical Engineering\/}, {\em 14\/},
  275--294.

\bibitem[{Scheutz et~al.(2013)Scheutz, Briggs, Cantrell, Krause, Williams, \&
  Veale}]{scheutzetal13irs}
Scheutz, M., Briggs, G., Cantrell, R., Krause, E., Williams, T., \& Veale, R.
  (2013).
\newblock Novel mechanisms for natural human-robot interactions in the {DIARC}
  architecture.
\newblock {\em Proceedings of {AAAI} Workshop on Intelligent Robotic
  Systems\/}.

\bibitem[{Scheutz et~al.(2017)Scheutz, Krause, Oosterveld, Frasca, \&
  Platt}]{scheutz2017spoken}
Scheutz, M., Krause, E., Oosterveld, B., Frasca, T., \& Platt, R. (2017).
\newblock Spoken instruction-based one-shot object and action learning in a
  cognitive robotic architecture.
\newblock {\em Proceedings of AAMAS\/}.

\bibitem[{Scheutz et~al.(2015)Scheutz, Malle, \& Briggs}]{scheutz2015towards}
Scheutz, M., Malle, B., \& Briggs, G. (2015).
\newblock Towards morally sensitive action selection for autonomous social
  robots.
\newblock {\em Proc. of RO-MAN\/}.

\bibitem[{Scheutz et~al.(2018)Scheutz, Williams, Krause, Oosterveld, Sarathy,
  \& Frasca}]{DIARC}
Scheutz, M., Williams, T., Krause, E., Oosterveld, B., Sarathy, V., \& Frasca,
  T. (2018).
\newblock An overview of the distributed integrated cognition affect and
  reflection {DIARC} architecture.
\newblock In {\em Cognitive architectures\/}.

\bibitem[{Searle(1975)}]{searle1975indirect}
Searle, J.~R. (1975).
\newblock Indirect speech acts.
\newblock {\em Syntax and Semantics\/}, {\em 3\/}, 59--82.

\bibitem[{Sharkey \& Sharkey(2010)}]{sharkey2010crying}
Sharkey, N., \& Sharkey, A. (2010).
\newblock The crying shame of robot nannies: an ethical appraisal.
\newblock {\em Interaction Studies\/}, {\em 11\/}, 161--190.

\bibitem[{Simmons et~al.(2011)Simmons, Makatchev, Kirby, Lee
  et~al.}]{simmons2011believable}
Simmons, R., Makatchev, M., Kirby, R., Lee, M.~K., et~al. (2011).
\newblock Believable robot characters.
\newblock {\em AI Magazine\/}.

\bibitem[{Tellex et~al.(2013)Tellex, Thaker, Deits, Simeonov, Kollar, \&
  Roy}]{tellex2013toward}
Tellex, S., Thaker, P., Deits, R., Simeonov, D., Kollar, T., \& Roy, N. (2013).
\newblock Toward information theoretic human-robot dialog.
\newblock {\em Robotics: Science and Systems\/}, {\em 32\/}, 409--417.

\bibitem[{Trott \& Bergen(2017)}]{trott2017theoretical}
Trott, S., \& Bergen, B. (2017).
\newblock A theoretical model of indirect request comprehension.
\newblock {\em Proceedings of the AAAI Fall Symposium Series on Artificial
  Intelligence for Human-Robot Interaction\/}.

\bibitem[{Verbeek(2011)}]{verbeek2011moralizing}
Verbeek, P.-P. (2011).
\newblock {\em Moralizing technology: Understanding and designing the morality
  of things\/}.
\newblock University of Chicago Press.

\bibitem[{Wada \& Shibata(2007)}]{wada2007living}
Wada, K., \& Shibata, T. (2007).
\newblock Living with seal robots -- its sociopsychological and physiological
  influences on the elderly at a care house.
\newblock {\em IEEE Transactions on Robotics\/}, {\em 23\/}, 972--980.

\bibitem[{Walker et~al.(2004)Walker, Lamere, Kwok, Raj, Singh, Gouvea, Wolf, \&
  Woelfel}]{walker2004sphinx}
Walker, W., Lamere, P., Kwok, P., Raj, B., Singh, R., Gouvea, E., Wolf, P., \&
  Woelfel, J. (2004).
\newblock Sphinx-4: A flexible open source framework for speech recognition.

\bibitem[{Wen et~al.(2018)Wen, Stewart, Billinghurst, Dey, Tossell, \&
  Finomore}]{wen2018hesitates}
Wen, J., Stewart, A., Billinghurst, M., Dey, A., Tossell, C., \& Finomore, V.
  (2018).
\newblock He who hesitates is lost (...in thoughts over a robot).
\newblock {\em Proceedings of the Technology, Mind, and Society\/}.

\bibitem[{Williams(2017)}]{williams2017iib}
Williams, T. (2017).
\newblock A consultant framework for natural language processing in integrated
  robot architectures.
\newblock {\em IEEE Intelligent Informatics Bulletin\/}.

\bibitem[{Williams et~al.(2016)Williams, Acharya, Schreitter, \&
  Scheutz}]{williams2016hri}
Williams, T., Acharya, S., Schreitter, S., \& Scheutz, M. (2016).
\newblock Situated open world reference resolution for human-robot dialogue.
\newblock {\em Proceedings of HRI\/}.

\bibitem[{Williams et~al.(2015)Williams, Briggs, Oosterveld, \&
  Scheutz}]{williams2015aaai}
Williams, T., Briggs, G., Oosterveld, B., \& Scheutz, M. (2015).
\newblock Going beyond command-based instructions: Extending robotic natural
  language interaction capabilities.
\newblock {\em Proceedings of AAAI\/}.

\bibitem[{Williams et~al.(2018{\natexlab{a}})Williams, Jackson, \&
  Lockshin}]{williams2018cogsci}
Williams, T., Jackson, R.~B., \& Lockshin, J. (2018{\natexlab{a}}).
\newblock A bayesian analysis of moral norm malleability during clarification
  dialogues.
\newblock {\em Proceedings of COGSCI\/}. Cognitive Science Society.

\bibitem[{Williams \& Scheutz(2016)}]{williams2016aaai}
Williams, T., \& Scheutz, M. (2016).
\newblock A framework for resolving open-world referential expressions in
  distributed heterogeneous knowledge bases.
\newblock {\em Proceedings of AAAI\/}.

\bibitem[{Williams \& Scheutz(2018)}]{williams2018oxford}
Williams, T., \& Scheutz, M. (2018).
\newblock Reference in robotics: A givenness hierarchy theoretic approach.
\newblock In J.~Gundel \& B.~Abbott (Eds.), {\em The oxford handbook of
  reference\/}.

\bibitem[{Williams et~al.(2018{\natexlab{b}})Williams, Thames, Novakoff, \&
  Scheutz}]{williams2018hri}
Williams, T., Thames, D., Novakoff, J., \& Scheutz, M. (2018{\natexlab{b}}).
\newblock {``T}hank you for sharing that interesting fact!'': Effects of
  capability and context on indirect speech act use in task-based human-robot
  dialogue.
\newblock {\em Proceedings of HRI\/}.

\bibitem[{Williams et~al.(2018{\natexlab{c}})Williams, Yazdani, Suresh,
  Scheutz, \& Beetz}]{williams2018auro}
Williams, T., Yazdani, F., Suresh, P., Scheutz, M., \& Beetz, M.
  (2018{\natexlab{c}}).
\newblock Dempster-shafer theoretic resolution of referential ambiguity.
\newblock {\em Autonomous Robots\/}.

\end{thebibliography}

}


\end{document}